\documentclass[letterpaper, 10pt, conference]{ieeeconf}  % Comment this line out
\usepackage{amssymb,graphicx,amsmath,color,algorithm,algorithmic}
\usepackage[noadjust]{cite} % groups citations
\usepackage[english]{babel}
\usepackage{blindtext}

\DeclareMathOperator*{\argmax}{arg\,max}

\IEEEoverridecommandlockouts                              % This command is only
\title{\LARGE \bf
TacWhiskers:
Biomimetic optical tactile whiskered robots
\vspace{0em}
}

\author{Nathan F. Lepora, Martin Pearson, Luke Cramphorn % <-this % stops a space
\thanks{NL was supported in part by a grant from the Leverhulme Trust on `A biomimetic forebrain for robot touch' (RL-2016-39).}% <-this % stops a space
\thanks{All authors are with the Department of Engineering Mathematics and Bristol Robotics Laboratory, University of Bristol, Bristol, UK.}
\thanks{Email: n.lepora@bristol.ac.uk}
}

\begin{document}

\maketitle
\thispagestyle{empty}
\pagestyle{empty}

% keywords: 	Force and Tactile Sensing 	Biomimetics

%%%%%%%%%%%%%%%%%%%%%%%%%%%%%%%%%%%%%%%%%%%%%%%%%%%%%%%%%%%%%%%%%%%%%%%%%%%%%%%%
\begin{abstract}
Here we propose and investigate a novel vibrissal tactile sensor - the TacWhisker array - based on modifying a 3D-printed optical cutaneous (fingertip) tactile sensor - the TacTip. Two versions are considered: a static TacWhisker array analogous to immotile tactile vibrissae (e.g. rodent microvibrissae) and a dynamic TacWhisker array analogous to motile tactile vibrissae (e.g. rodent macrovibrissae). Performance is assessed on an active object localization task. The whisking motion of the dynamic TacWhisker leads to millimetre-scale location perception, whereas perception with the static TacWhisker array is relatively poor when making dabbing contacts. The dynamic sensor output is dominated by a self-generated motion signal, which can be compensated by comparing to a reference signal. Overall, the TacWhisker arrays give a new class of tactile whiskered robots that benefit from being relatively inexpensive and customizable. Furthermore, the biomimetic basis for the TacWhiskers fits well with building an embodied model of the rodent sensory system for investigating animal perception. 
\vspace{-0em}
\end{abstract}

%The TacWhisker sensor holds promise for applications of whiskers and embodied neurorobotic investigations of tactile vibrissal sensing in rodents.e.  

%%%%%%%%%%%%%%%%%%%%%%%%%%%%%%%%%%%%%%%%%%%%%%%%%%%%%%%%%%%%%%%%%%%%%%%%%%%%%%%%
\section{INTRODUCTION}

Animal sensory organs have inspired diverse artificial sensing systems, from chemosensing based on olfaction to whiskered robots based on the vibrissal sense of rodents. We are intimately familiar with our own human sense of touch, which underlies our abilities to manipulate the world and interact with other human beings. Yet for rodents and some other mammals, their primary tactile sense is from vibrissae, or tactile hair, which functions as a proximity sense for navigation and to catch prey. Whiskered robots are of interest to deploy this proximity sense in applications where vision is compromised, such as disaster recovery, and as biomimetic instantiations to investigate the physiological and neuroscientific principles underlying natural sensing~\cite{Prescott2009}. 

Accordingly, whiskered robots are an active area of research, with state-of-the-art devices including the SHREWbot and BELLAbot robots~\cite{Pearson2011,Assaf2016a}. While these whiskered robots are excellent biomimetic counterparts of biological vibrissal systems, they are highly complex, one-off robots that are expensive and labour-intensive to make; for example, the SHREWbot nose cone has 24 whisker modules, each having an actuated base with a Hall effect sensor to measure whisker deflection. Meanwhile, within the related area of artificial cutaneous (fingertip) touch, there has been progress towards 3D-printed, optical tactile sensors such as the TacTip family~\cite{Chorley2009,Ward-Cherrier2018}, which are inexpensive and simple to fabricate. 

\begin{figure}[t!]
	\centering
	\includegraphics[width=0.46\columnwidth,trim={0 30 0 30},clip]{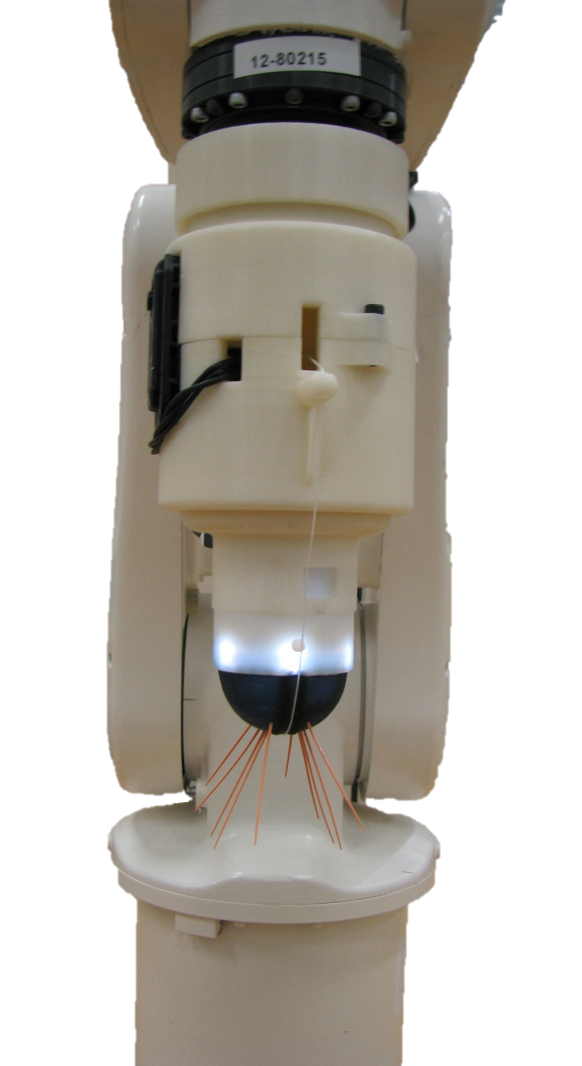}
	\includegraphics[width=0.45\columnwidth,trim={0 -50 0 0},clip]{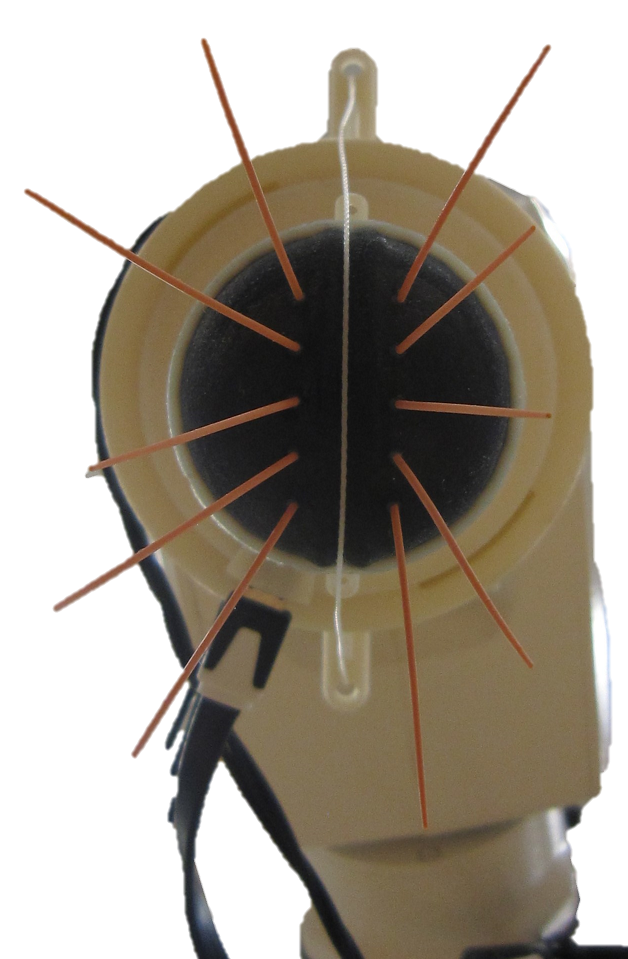}\vspace{0em}
	\caption{Side (left image) and front (right image) views of the dynamic TacWhisker array mounted on an ABB robot arm. The actuation module, sensor body and tip are visible, with the tendon that protracts the whiskers.}
	\label{fig:1}
	\vspace{-1em}
\end{figure}

Here we propose a new class of 3D-printed, optical tactile whisker sensors that we call TacWhiskers, based on the TacTip design. We consider two versions: a static TacWhisker array analogous to immotile tactile vibrissae ({\em e.g.} rodent microvibrissae) and a dynamic TacWhisker array analogous to motile tactile vibrissae ({\em e.g.} rodent macrovibrissae). The dynamic TacWhisker uses a single motor to protract and retract its whiskers in a rhythmic whisking motion, using a tendon passing between two rows of whiskers (Fig.~\ref{fig:1}).

To assess the TacWhisker performance, we consider a localization task where the position of a rod is classified from whisker deflections, analogously to an experimental protocol for rodent perception~\cite{Diamond2008}. We find the accuracy of TacWhisker perception depends strongly on the whisker motion: the static TacWhisker has poor location perception with a forward/back dabbing motion and the dynamic TacWhisker has good location perception when whisking. These motions are then applied to an active perception task, in which the rod is both localized and centred within the whisker array. Only the dynamic TacWhisker is able to perform well at this task, with trajectories quickly centring on the stimulus and localization performance near perfect after a few contacts. 

%%%%%%%%%%%%%%%%%%%%%%%%%%%%%%%%%%%%%%%%%%%%%%%%%%%%%%%%%%%%%%%%%%%%%%%%%%%%%%%%

\begin{figure}[t]
	\centering
	\includegraphics[width=0.9\columnwidth,trim={0 0 0 0},clip]{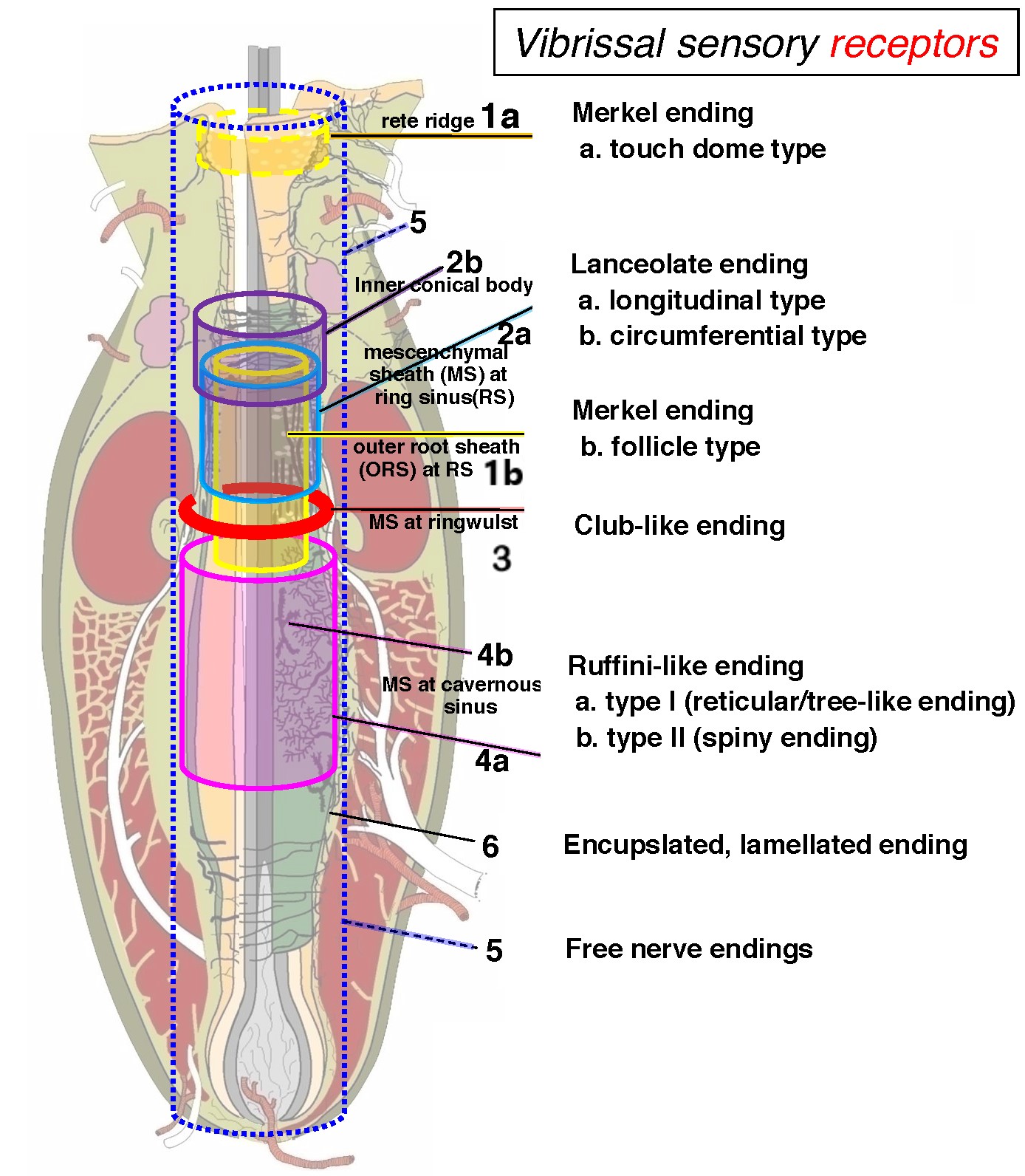}
	\caption{Physiology of tactile vibrissae - the follicle-sinus complex from which the whisker shaft emanates, and within which are the vibrissal sensory mechanoreceptors. (Diagram modified from~\cite{Ebara2017} under a Creative Commons Attribution-ShareAlike 3.0 License.) }
	\label{fig:2}
	\vspace{-1em}
\end{figure}

\section{BACKGROUND AND RELATED WORK}

\subsection{Biological sensing with tactile hair}

Almost all mammals except {\em Homo Sapiens} have a specialised form of tactile hair called vibrissae  or whiskers~\cite{Ahl1986}. Whiskers differ from conventional (pelagic) hair by being~\cite{Ahl1986}: (i) much longer; (ii) primarily facial (although they can also occur on the body); (iii) sited on a large, highly-innervated follicle-sinus complex (Fig.~\ref{fig:2}); and (iv)~specifically represented in the sensory cortex of the brain.  

Tactile whiskers may be motile or immotile, depending on the animal and where on the body they are located~\cite{Prescott2011,Ahl1986}. In mice and rats, the long facial whiskers (macrovibrissae) around the snout move bilaterally back and forth in an active sensing motion known as `whisking'; meanwhile, the short facial whiskers (microvibrissae) underneath the nostrils are fixed. Immotile whiskers are also found on other body regions; for example, the carpal vibrissae just above cat paws. 

Sensory mechanoreceptors within the whisker follicle transduce motion of the whisker shaft into contact information about the environment~\cite{Ebara2017}. Merkel cells in a collar around the follicle opening activate slowly-adapting (SA) neurons that fire during sustained whisker deformation. Deeper receptors such as Lanceloate, Ruffini and Paccinian endings within the whisker follicle act in concert with the Merkel cells, signalling information about vibration, motion and whisker deflection, to comprise the vibrissal tactile sense. 

\subsection{Biomimetic tactile whiskered robots}

Over the last decade there have been a succession of biomimetic tactile whiskered robots developed from a collaboration between Sheffield Robotics and Bristol Robotics Laboratory~\cite{Prescott2009}. The initial Whiskerbot mobile platform had 6 glass-fibre moulded whiskers mounted on strain gauges to measure 2D deflections of the whisker shaft~\cite{Pearson2007}. An improved SCRATCHbot platform had 18 actuated 3D-printed whiskers with Hall effect sensors to measure deflections while actively whisking~\cite{Pearson2010}. These single-actuated whiskers were modularized as part of the BIOTACT project, leading to another mobile whiskered platform called Shrewbot~\cite{Pearson2011} and a stand-alone whisker array for mounting on a robot arm~\cite{Sullivan2012}, both with 24 individually-actuated whiskers from $\sim$5-15\,cm long arranged in a conical 6-by-4 array. 

Other technologies have been used for whiskered robots. An early mobile robot from the aMouse project attached real rodent whiskers to microphones for detecting high-frequency vibrations~\cite{Fend2005}. Soon after, a whisker array was built with both slowly adapting (deflection) and rapidly adapting (velocity) components of the whisker signal and used to recognize shape~\cite{Kim2007}. A more recent robot, BELLAbot, combined the BIOTACT technology with advances in electroactive polymeric (EAP) actuation to orient, whisk and sense over an array of 20 distinct EAP whisker modules~\cite{Assaf2016a}. Simpler biomimetic whiskers have also been proposed, such as using strain gauges in the follicle to measure the bending moments proposed to underlie localization with rodent vibrissae~\cite{Emnett2018}.
 
\subsection{Biomimetic optical tactile sensors}

\begin{figure}[t!]
	\centering
	\begin{tabular}[b]{@{}cc@{}}
		\small\bfseries{(a) TacTip transduction}&
		\small\bfseries{(b) TacWhisker transduction}\\
		\includegraphics[width=0.5\columnwidth,trim={0 0 0 0},clip]{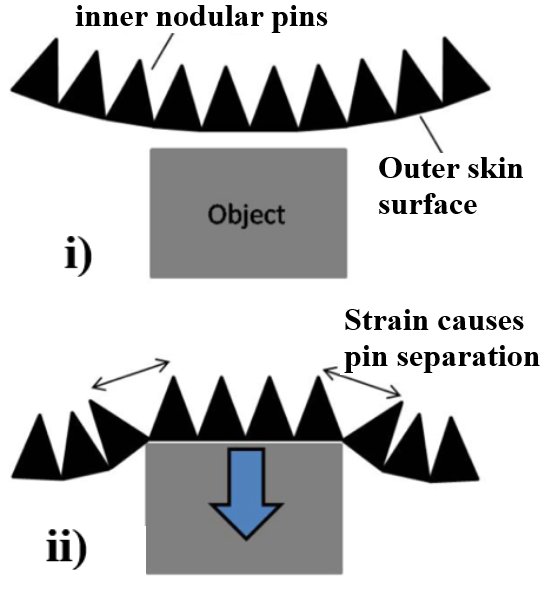}&
		\includegraphics[width=0.38\columnwidth,trim={0 0 0 0},clip]{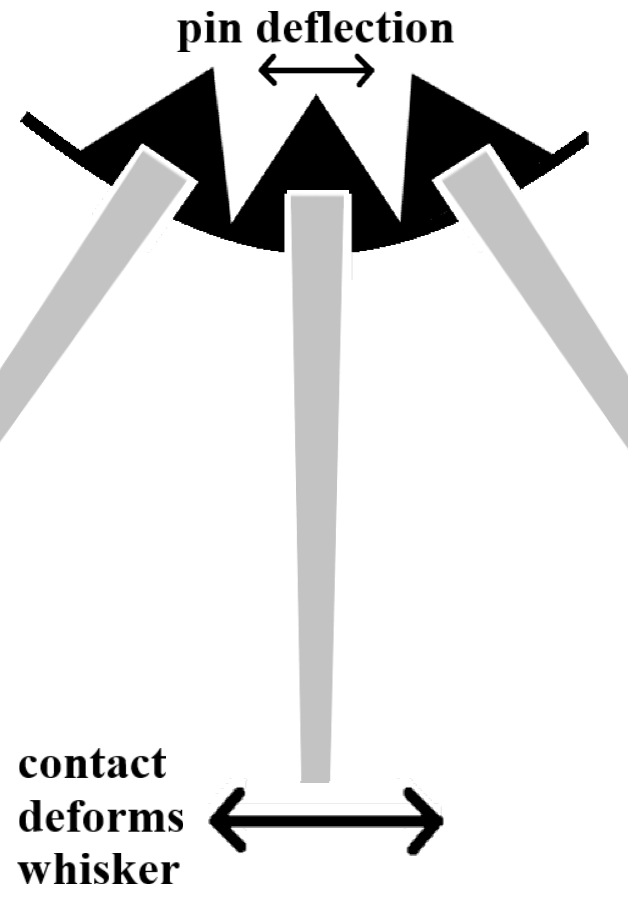}
	\end{tabular}
	\caption{Common transduction principle of the TacTip and TacWhiskers. (a)~For the TacTip, surface strain separates the inner pins. (b) For the TacWhisker, whisker shaft deformation deflects the pins. In both cases, the movement of markers on the pin tips is tracked by a camera.}
	\label{fig:3}
	\vspace{-1em}
\end{figure}

This paper investigates a novel vibrissal tactile sensor based on modifying a 3D-printed cutaneous (fingertip) tactile sensor called the TacTip (see~\cite{Ward-Cherrier2018} for a recent overview).   

The TacTip is a biomimetic tactile sensor based on the layered structure of human glabrous skin~\cite{Chorley2009}. It has an outer biomimetic epidermis made from a rubber-like material over an inner biomimetic dermis made from polymer gel. These two materials interdigitate in a mesh of inner nodular pins, based on the intermediate ridge structure of human skin that extends under the epidermis into the dermis. The biomimetic counterparts to sensory mechanoreceptors are markers on the pin tips, which can be imaged through a transparent gel that comprises the dermis. The pin movement is considered analogous to Merkel cell activity in the intermediate ridges~\cite{Cramphorn2017}.

A principal observation underlying this paper is that the transduction mechanism in the TacTip can also be applied to tactile whiskers (Fig~\ref{fig:3}). Information about how the TacTip surface deforms upon contacting an object is represented in the movement of markers on the internal pins, {\em e.g.} separating on regions of high spatial curvature (Fig.~\ref{fig:3}a). Instead, for internal pins attached to whiskers extending out of the sensor surface, the markers represent displacement of the whisker shafts (Fig.~\ref{fig:3}b). As the same receptor type (Merkel cells) is implicated in both types of biological sensing, this mechanism also gives a commonality in the biomimetics of cutaneous and vibrissal touch (although it should be noted that a multitude of other mechanoreceptor types are also involved in both types of touch). 

%%%%%%%%%%%%%%%%%%%%%%%%%%%%%%%%%%%%%%%%%%%%%%%%%%%%%%%%%%%%%%%%%%%%%%%%%%%%%%%%

\begin{figure}[t]
	\centering
	\begin{tabular}[b]{@{}cc@{}}
		\begin{tabular}[b]{@{}c@{}}
			\small\bfseries{(a) CAD for the whisker}\\
			\includegraphics[width=0.4\columnwidth,trim={0 75 0 35},clip]{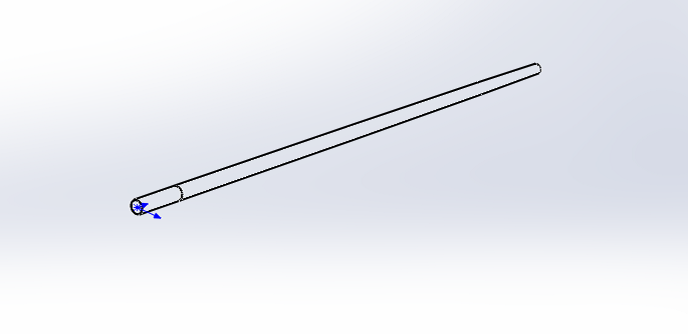}\\
			\small\bfseries{(b) CAD for the tip}\\
			\includegraphics[width=0.32\columnwidth,trim={0 40 0 20},clip]{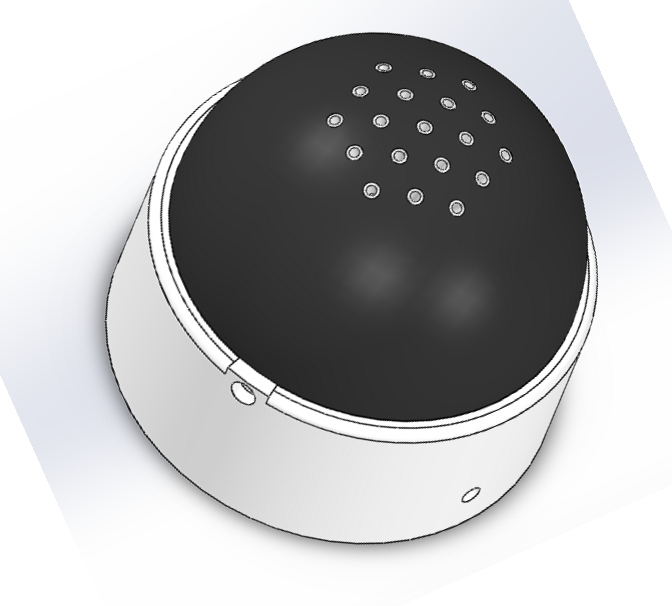}\\
			\small\bfseries{(c) CAD for the base}\\
			\includegraphics[width=0.4\columnwidth,trim={0 5 0 5},clip]{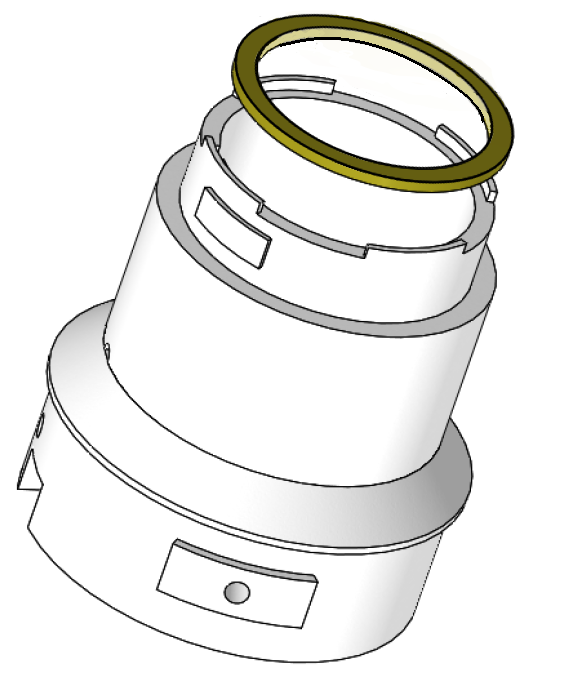}\\
		\end{tabular}&
		\begin{tabular}[b]{@{}c@{}}
			\small\bfseries{(d) Static TacWhisker array}\\
			\includegraphics[width=0.49\columnwidth,trim={195 10 235 10},clip]{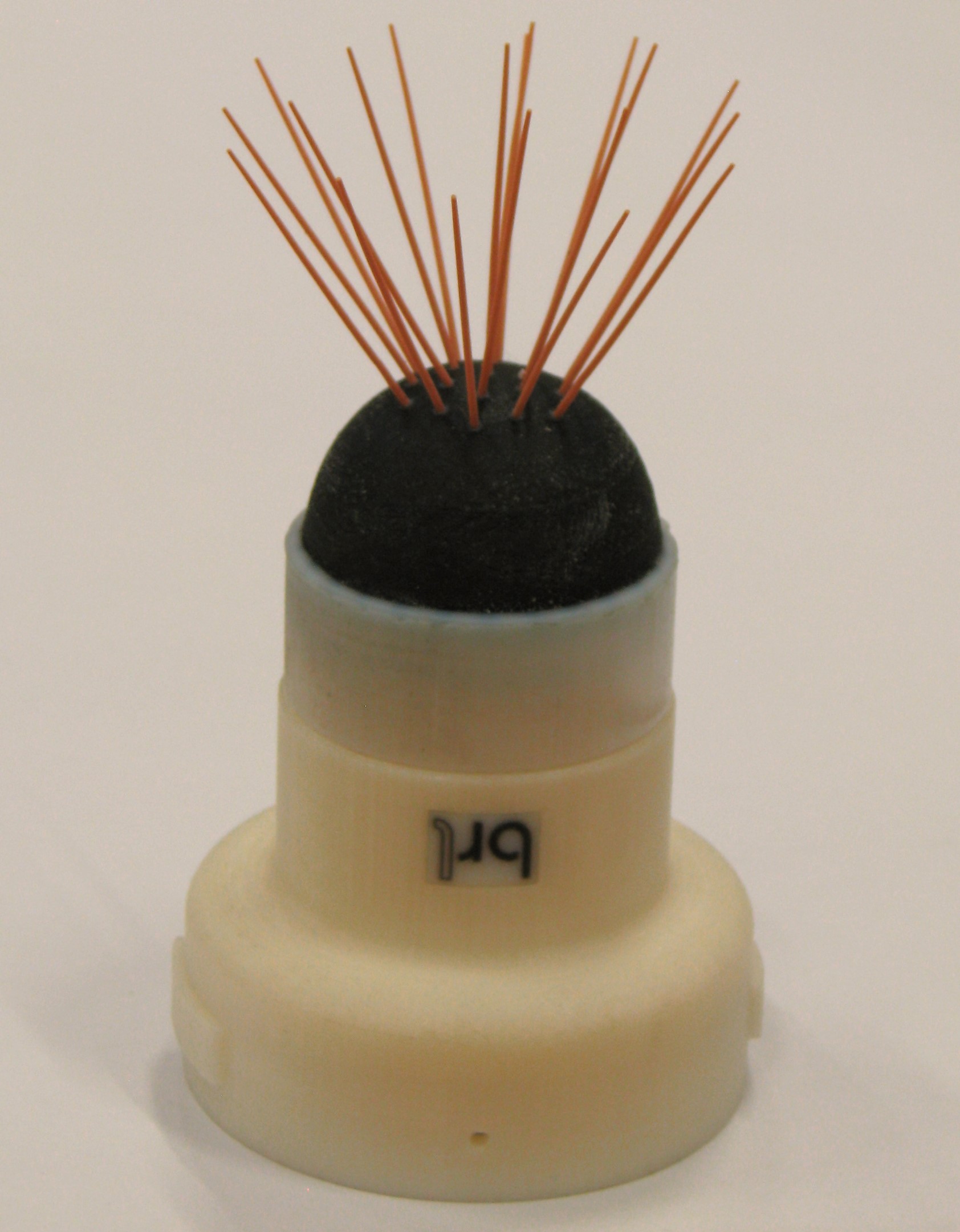}
		\end{tabular}
	\end{tabular}
	\caption{Design of the static TacWhisker array. (a) CAD for the 3D-printed whisker, a tapered shaft. (b) CAD for the tip, comprising a compliant skin with sockets for the whiskers and a rigid base. (c) CAD for the base housing for the camera, showing also the LED ring. (d) Assembled TacWhisker array. }
	\label{fig:4}
	\vspace{-1em}
\end{figure}

\begin{figure}[t!]
	\centering
	\includegraphics[width=0.8\columnwidth,trim={50 10 30 30},clip]{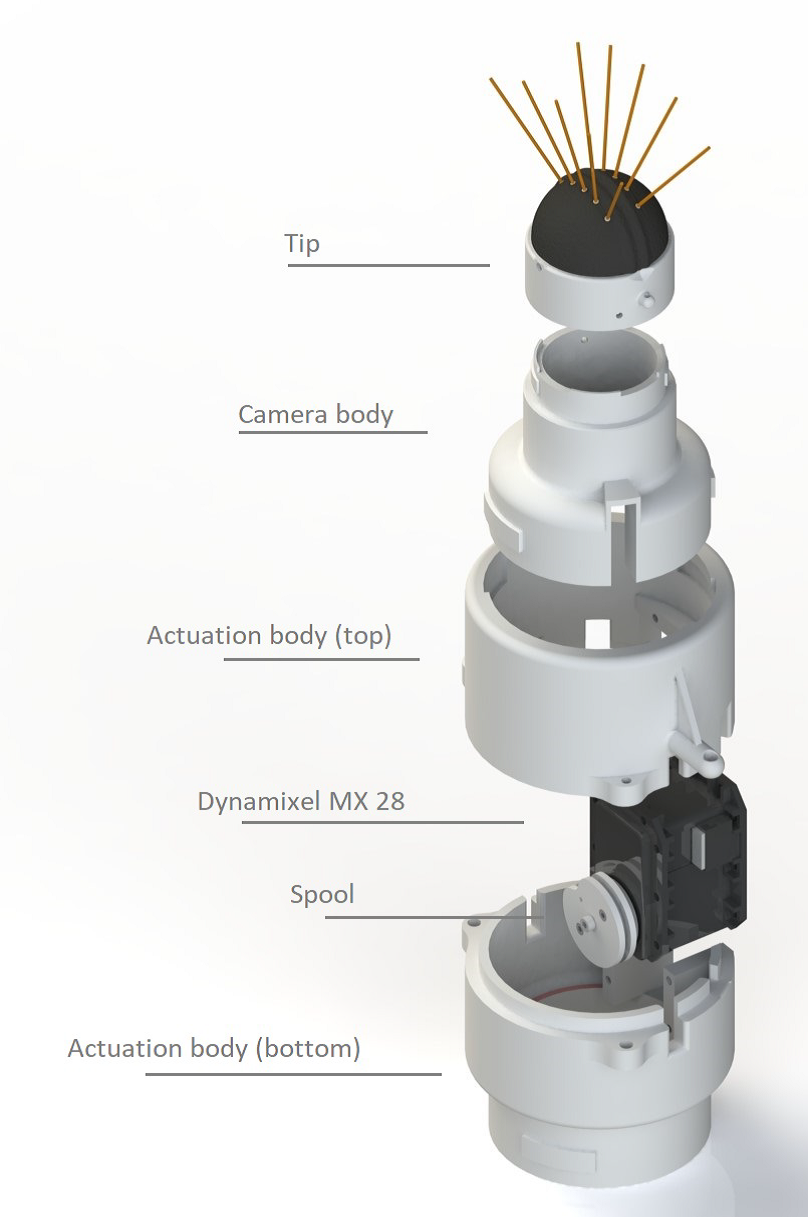}
	\caption{Design of the dynamic TacWhisker array. The 3D-printed tip with mounted whiskers attaches to the base housing the camera, which attaches to the actuation module comprising the motor and housing. A tendon runs from the spool, through guides and across a groove in the compliant tip; actuation of the tendon causes the tip to deform, moving the whiskers.}
	\label{fig:5}
	\vspace{1em}
	\begin{tabular}[b]{@{}cc@{}}
		\small\bfseries{(a) Active protraction}&
		\small\bfseries{(b) Passive retraction}\\
		\includegraphics[width=0.4\columnwidth,trim={1200 120 200 600},clip]{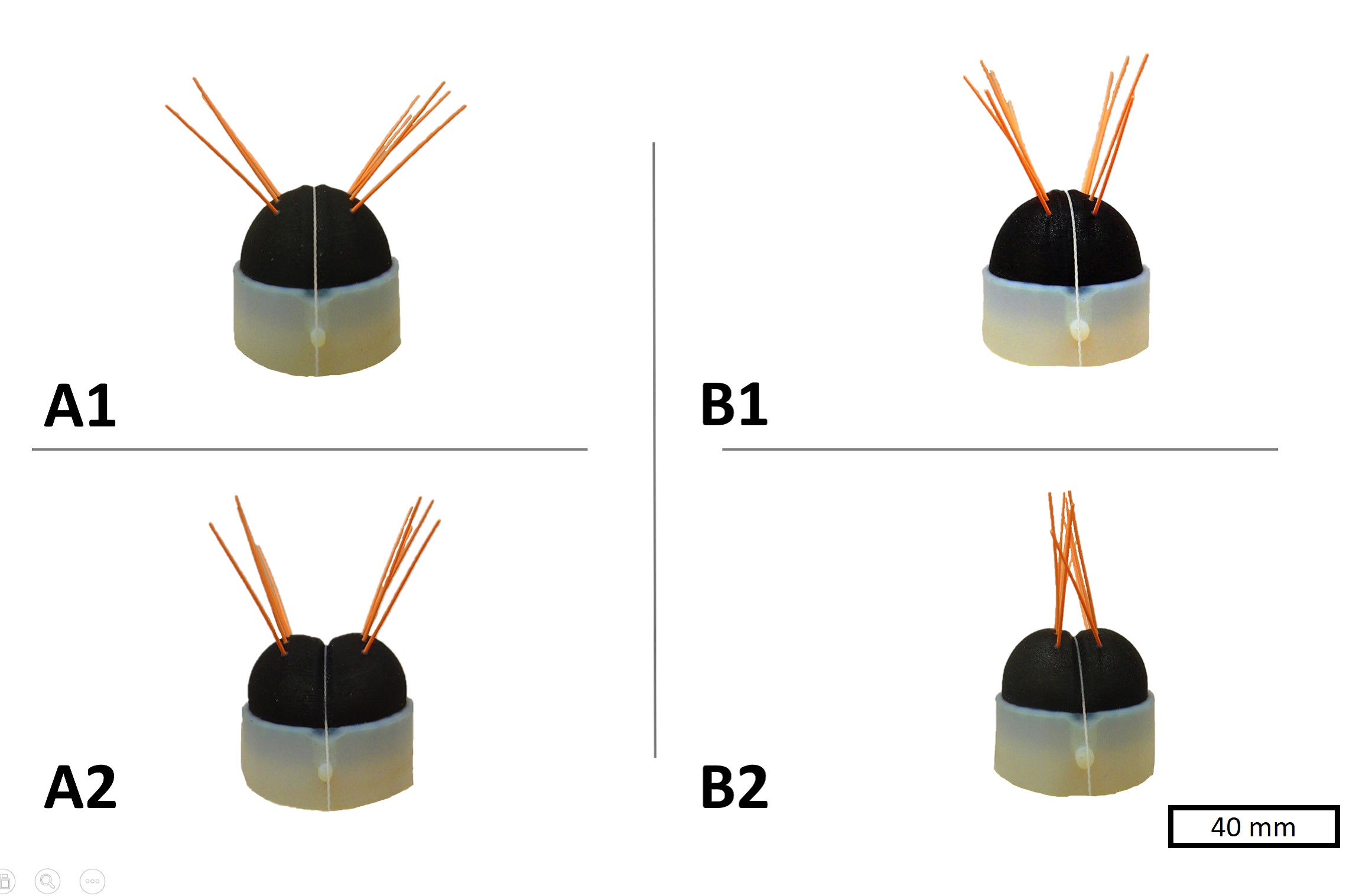}&		\includegraphics[width=0.4\columnwidth,trim={1200 670 200 50},clip]{fig6_comp}
	\end{tabular}
	\caption{Whisking motion of the dynamic TacWhisker array. (a) The motor pulls on the tendon to actively protract (bring together) the two rows of whiskers. (b) Reversing the motor releases the tendon to passively retract (pull apart) the whiskers by elastic reformation of the tip.}
	\label{fig:6}
	\vspace{-1em}
\end{figure}

\section{METHODS}
\label{sec:3}

\subsection{Inspiration and design of the TacWhisker}
\label{sec:3a}

We call the whiskered version of the TacTip a {\em TacWhisker} array, emphasising it is based on tactile whiskers rather than tactile (finger)tips. In this work, we consider two types of TacWhisker array: static (immotile) and dynamic (motile).

\begin{figure*}[h!]
	\centering
	\includegraphics[width=0.24\textwidth,trim={-20 5 -60 0},clip]{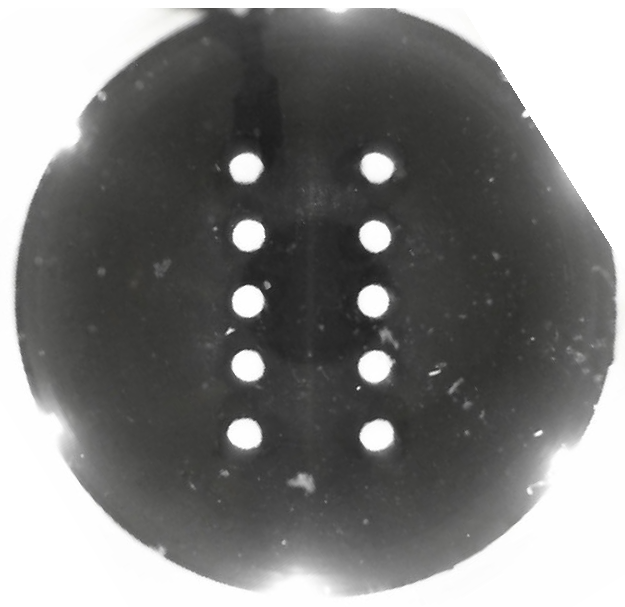}
	\includegraphics[width=0.24\textwidth,trim={-30 -5 -50 0},clip]{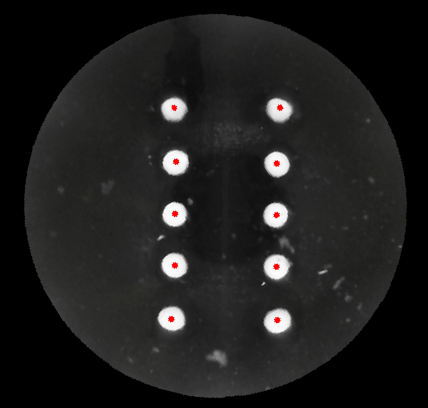}
	\includegraphics[width=0.24\textwidth,trim={-10 10 -20 0},clip]{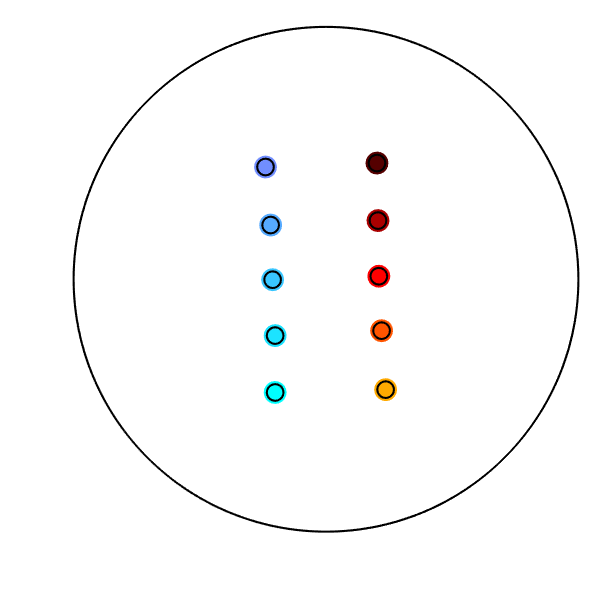}
	\includegraphics[width=0.23\textwidth,trim={-0 0 -20 0},clip]{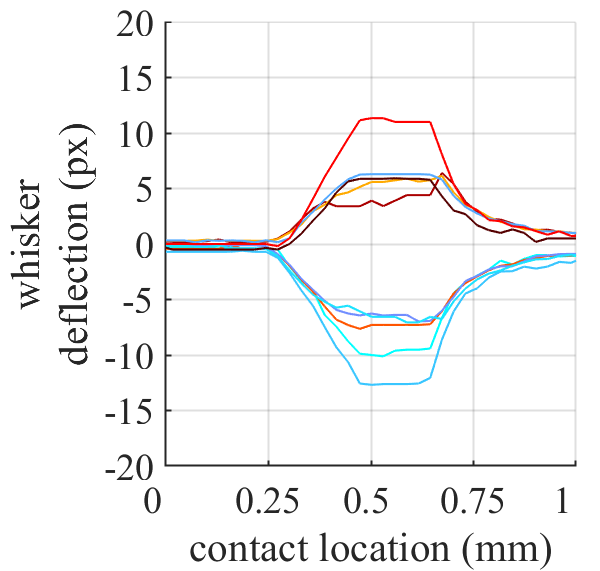}
	\includegraphics[width=\textwidth,trim={-10 20 -10 10},clip]{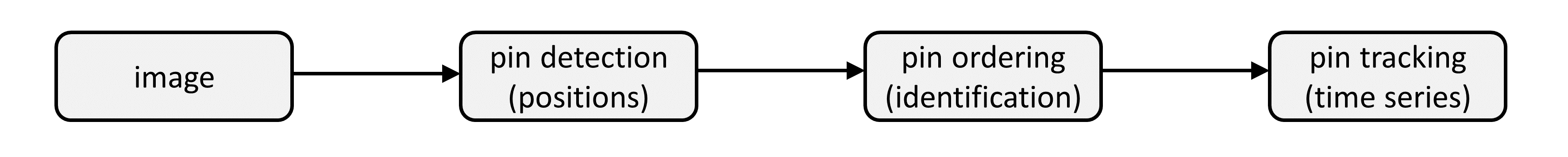}
	\caption{Data processing pipeline. The internal camera captures an image of the pins attached to the shafts of the whiskers. The pins are detected and located with a blob detection algorithm. The pins are then ordered, by tracking the pins from frame-to-frame (here coloured by their row). Over multiple frames, the processing pipeline produces a time series of pin movement, representing deflection of the whiskers.}
	\label{fig:7}
	\vspace{-1em}
\end{figure*}

\subsubsection{Static TacWhisker array (Fig.~\ref{fig:4})}

The first design of TacWhisker array modifies just the standard TacTip tip to house whiskers (Figs~\ref{fig:4}a,b). There is no modification of the 3D-printed TacTip base (Fig.~\ref{fig:4}c), which contains the USB camera (Lifecam, Microsoft) and an LED ring to illuminate the pin markers (see~\cite{Ward-Cherrier2018} for details). The tip is based on recent versions of the TacTip~\cite{Ward-Cherrier2018} that use multi-material 3D-printing. The compliant surface and inner pins printed in a rubber-like material (Tango Black+ 27) and the pin tips and mount in hard plastic (Vero White); this outer surface is filled with a soft clear silicone gel (Techsil RTV27905) held in place with a clear acrylic lens cap. 

For housing whiskers, the tip is modified to (Fig.~\ref{fig:4}b): (i)~reduce the number of pins to 21 (from 127) sited near the top of the tip; (ii)~space the pins further apart (4.5\,mm separation, rather than 3\,mm keeping the hexagonal projection); (iii)~enlarge and extrude the solid markers through the compliant surface (2.2\,mm dia.$\times$3.5\,mm depth pins, increased from 1.2\,mm$\times$2\,mm); and (iv)~include a hole (1\,mm dia.$\times$3\,mm depth) functioning as a socket for the whiskers. These design choices were chosen to give good pin movement upon deflection of the whiskers, and to site the whiskers appropriately for contact. 

The whiskers~(Fig.~\ref{fig:4}a) are modified versions of BIOTACT vibrissae~\cite{Pearson2011} that are 3D printed using nanocure-25. The main change is to reduce the whisker size for the smaller scale of the TacTip (40\,mm dia.) compared with the BIOTACT conical housing (100\,mm dia.). Accordingly, we chose whiskers 40\,mm long with a 0.98\,mm dia. base tapering to 0.6\,mm dia. at the tip, similar in scale to real rat whiskers. For simplicity, all whiskers had the same dimensions, but it would be straightforward to introduce size variations like those of rodent macrovibrissae.

\subsubsection{Dynamic TacWhisker array (Fig.~\ref{fig:5})}

The TacWhisker array can join onto an actuation module that protracts and retracts the whiskers back and forth in a whisking motion (Fig.~\ref{fig:6}). The whiskered tip is modified to have 2 rows of 5 whiskers arranged in a bilaterally symmetric pattern. A tendon runs through a groove between these rows and two guides in the tip mount (Fig.~\ref{fig:6}). Forwards whisker motion (protraction) results from tensioning the tendon to compress the surface at the midline (Fig.~\ref{fig:6}a); backwards whisker motion (retraction) results from releasing the tendon to elastically reform the surface (Fig.~\ref{fig:6}b). The compliant surface and whisker mounts are shaped so that the whisker tips can meet under modest surface compression. The dynamic TacWhisker array can thus rhythmically protract and retract its whiskers together and apart in a motion akin to rodent whisking.

The dynamic TacWhisker is designed to be modular and re-use parts of the static TacWhisker. Apart from the modified whiskered tip, the TacWhisker base housing the camera and LED lighting is the same as the conventional TacTip.  The underside of the base has a bayonet fitting, which is used to connect to an actuation module for driving the tendon (Fig.~\ref{fig:5}). This actuation module houses a Dynamixel MX 28 servomotor and spool for the tendon, with outer guides to ensure the tendon runs smoothly from the spool, outside the actuation and body modules, and over the TacWhisker tip. %The base of the actuation module has a bayonet fitting to mount similarly to the TacTip and static TacWhisker array.  

\subsection{Robotic platform and software infrastructure}
\label{sec:3b}

For testing, the static or dynamic TacWhisker array is mounted as an end-effector on a 6-DOF robotic arm (IRB 120, ABB) ({\em e.g.} Fig.~\ref{fig:1}). The arm is mounted on a table that also contains mounting stations for the stimuli. A custom 3D-printed mount is bolted to the rotating (wrist) section of the arm to which either sensor can be attached via a common bayonet fitting on the TacWhisker base and actuation module.

A modular software infrastructure is used in which MATLAB is the primary interface for running tests and analysing data. The ABB arm is controlled via an IronPython and RAPID interface, and data gathered from the USB camera within the TacWhisker sensor with Python OpenCV. Similarly, a Python interface controls the dynamixel motor of the dynamic TacWhisker array. Communication between software modules is via TCP/IP ports and sockets. 

\subsection{Sensory transduction and data processing}
\label{sec:3c}

\subsubsection{Sensing}

Following recent studies with the TacTip~\cite{Lepora2015,Ward-Cherrier2018}, the sensor output is a time series of pin deflections extracted from the camera images. The transformation of the camera image to marker positions requires that the pin markers be detected, which is done via standard `blob detection' methods in Python OpenCV. Overall, the data processing is a pipeline: camera image to pin detection to pin identification (nearest neighbour tracking) to give an ordered time series of pin deflections measured in pixels~(Fig.~\ref{fig:7}).

The resulting tactile whisker data comprises a multi-dimensional time series of $(x,y)$ pin deflections, measured in pixels on the horizontal and vertical directions of the camera image. For visualization, the time series plots of the $x$- and $y$-deflections are labelled by colouring the tactile dimension by its pin location (Fig.~\ref{fig:7}, right plots).

\begin{figure*}[t!]
	\centering
	\begin{tabular}[b]{@{}c@{}c@{}}
		\multicolumn{2}{c}{\bf (a) Static TacWhisker array with dabbing motion} \\
		\includegraphics[width=0.8\textwidth,trim={0 25 0 0},clip]{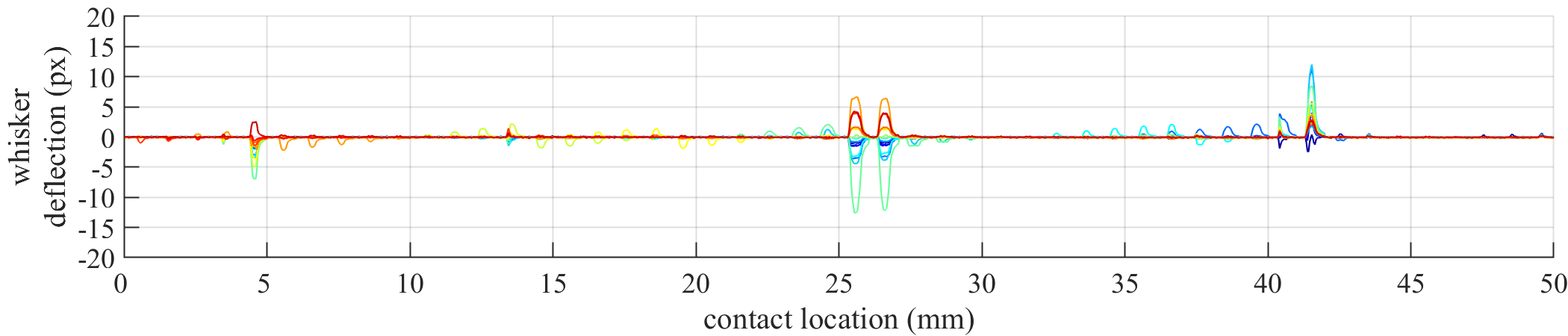} & 
		\includegraphics[width=0.16\textwidth,trim={-5 15 -5 0},clip]{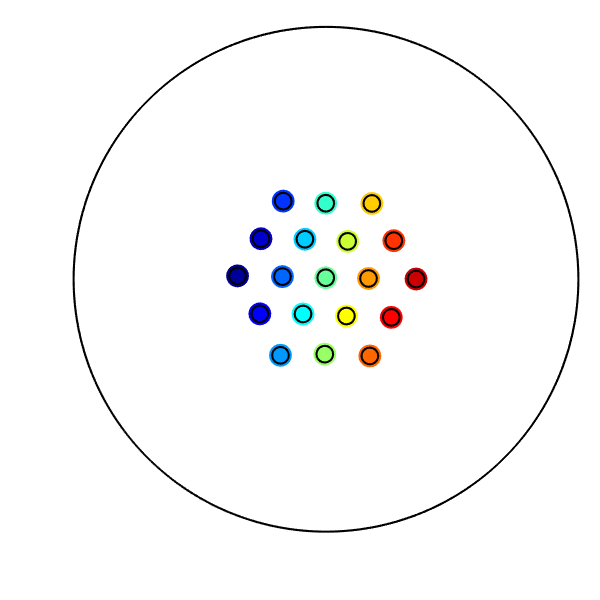} \\
		\multicolumn{2}{c}{\bf (b) Dynamic TacWhisker array with whisking motion} \\
		\includegraphics[width=0.8\textwidth,trim={0 25 0 0},clip]{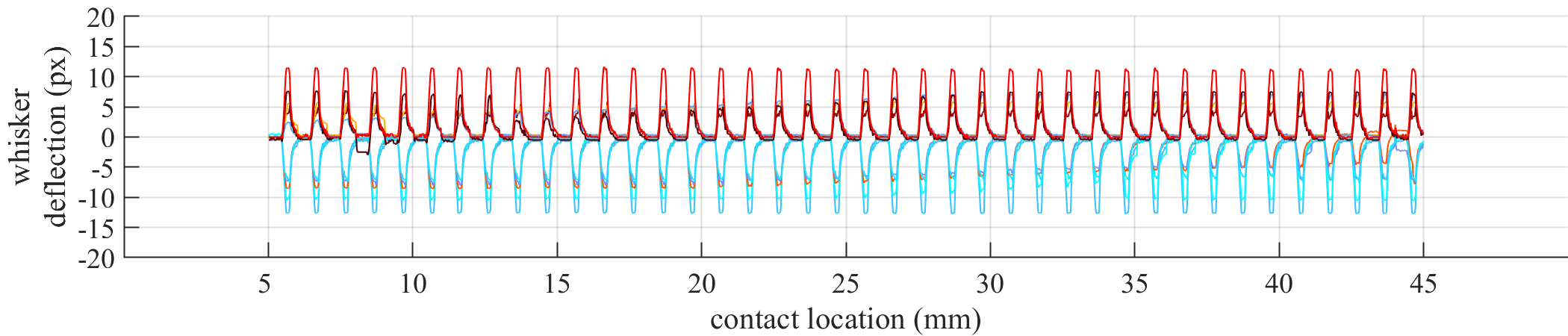} & 
		\includegraphics[width=0.16\textwidth,trim={-5 15 -5 0},clip]{fig8cr} \\
		\multicolumn{2}{c}{\bf (c) Dynamic TacWhisker array with self-motion calibration} \\
		\includegraphics[width=0.8\textwidth,trim={0 0 0 0},clip]{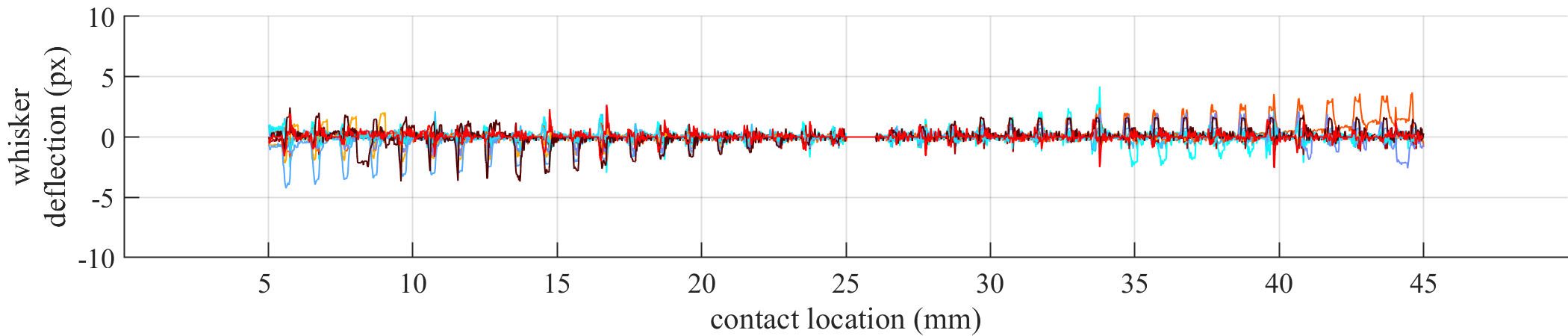} & 
		\includegraphics[width=0.16\textwidth,trim={-5 -20 -5 0},clip]{fig8cr} \\
	\end{tabular}
	\caption{Location data collected from the static (a) and dynamic (c,d) TacWhisker arrays. In all cases, the sensor is moved across a horizontal range from left to right (static: 50\,mm range; dynamic: 40\,mm range). The plot colour denotes the identity of the whiskers (right images).}
    \label{fig:8}
	\vspace{-1em}
\end{figure*}

\subsubsection{Perception}

Tactile perception is the process of inferring the properties of a stimulus from data collected by contacting that stimulus. Here we use a likelihood model that transforms tactile data $D$ into a likelihood probability $P(D\,|\,H_i)$ for a set of perceptual hypotheses $\{H_1,...,H_N\}$, which could be the labels ({\em e.g.} location $x_i$ in mm) for training data used to construct the model. The perceptual decision is then the hypothesis $H_i=\argmax_i P(D\,|\,H_i)$ that has the maximum likelihood for some sensed tactile data $D$.

Following recent studies with the TacTip, here we use a histogram likelihood model~\cite{Lepora2012,Lepora2016}, which bins the sensor data into intervals and counts bin frequency to form sampling distributions that are multiplied over sensor dimension and time. While simple, this model is effective for the TacTip and other sensors~\cite{Lepora2015,Lepora2016}, bears analogy with neural processing~\cite{Lepora2016} and is fairly robust and efficient. That said, the likelihood model is not the focus of this study, and so any model that works reasonably well would have been sufficient. 

All quantitative analyses of perceptual accuracy in this paper are based on cross validation over 10 repeated runs for data collection (representative single sets shown in Fig.~\ref{fig:8}). Monte Carlo sampling of a randomly-chosen training set and test data chosen from a different random set is then used. Typically 10,000 samples are used per analysis. 

\subsubsection{Active perception}

We follow the approach of `biomimetic active touch for fingertips and whiskers'~\cite{Lepora2016} in which active perceptual decisions are sequential over multiple tactile contacts $D(1),\cdots,D(T)$ with actions made between contacts to fulfil a goal, such as centring the sensor on a stimulus. Bayes rule is applied recursively to the likelihoods $P(D\,|\,H_i)$ to integrate evidence over contacts
$$P(H_i|D(t)) = \frac{P(D(t)|H_i)P(H_i|D(t-1))}{\sum_j P(D(t)|H_j)P(H_j|D(t-1))}$$
beginning from flat priors $P(H_i|D(0))=P(H_i)=1/N$. 

Here we use a simple active perception policy in which actions move the tactile sensor towards a goal location $x_{\rm fix}=H_{N/2}$, here taken as the stimulus centre. Then the actions are translations $\Delta x(t) = (x_{\rm fix}-x_j(t))$, with $x_j$ the $j$th location class $j = \argmax_i P(D(t)|H_i)$. The decision is made when the probability crosses a threshold $P(H_i|D(t))>\theta$ that sets how many contacts (on average) are needed for a decision.

%%%%%%%%%%%%%%%%%%%%%%%%%%%%%%%%%%%%%%%%%%%%%%%%%%%%%%%%%%%%%%%%%%%%%%%%%%%%%%%%

\section{RESULTS}

\subsection{Inspection and comparison of TacWhisker data}
\label{sec:4a}

Whisker contact data from two distinct experimental situations were collected (Figs~\ref{fig:8}). We chose motions in which the TacWhisker made discrete contacts with the stimulus, with the whiskers leaving the stimulus between contacts:\\ 
(a) Static TacWhisker array with dabbing motion. The sensor was moved horizontally across a rod keeping, dabbing vertically down (15\,mm) to make a tapping contact onto the rod stimulus, with 50 taps across 50\,mm.\\  
(b) Dynamic TacWhisker array with whisking motion. The experiment was repeated using the dynamic array to whisk onto the rod stimulus, with 40 whisks across 40\,mm. (The shorter range was due to to a narrower whisker field.)

A further dataset was created from modifying set (b):\\
(c) Dynamic TacWhisker array with self-motion calibration. A reference signal (taken at the centre of the location range) was subtracting from the whisking data; this calibration makes a whisker contact more visually apparent since the self-motion dominates otherwise.   

%Note that the change in whisker array morphology between the static and dynamic TacWhisker arrays ({\em c.f.} Fig.~\ref{fig:4} {\em vs} Fig.~\ref{fig:5}) gives different ranges for the sensor contacting the object (0-50\,mm in Figs~\ref{fig:8}a; 5-45\,mm in Figs~\ref{fig:8}b,c). 

In all experiments, good quality data were obtained from the TacWhisker arrays, as evident in the smoothly varying plots in Fig.~\ref{fig:8} with signal dominating over noise. Furthermore, the sensor data clearly covaries with contact location, which is especially evident in the dynamic array calibrated for self-motion (Fig.~\ref{fig:8}c), suggesting they will accurately perceive location. However, as we will explore below, the manner in which the whisker contacts the stimulus is key for perception, as is evident in the significant differences between the data in Figs~\ref{fig:8}a,b at the same locations.

\subsection{Self-motion affects the TacWhisker data}
\label{sec:4b}

A principal difference of the dynamic from the static TacWhisker arrays is that the tactile data is strongly affected by the self-motion of the dynamic array: the whiskers sweeping forwards then back is the most prominent feature of the data (Fig.~\ref{fig:8}b). This self-motion effect is also a known aspect of biological vibrissal systems (see discussion). Since the whisking motion remains constant frequency and amplitude in the present experimental task, its effect on the tactile data can be removed by subtracting a reference signal, which we choose to be at the centre of the location range. This gives the self-motion compensated tactile data (Fig.~\ref{fig:8}c), where the changing contact with the stimulus is clearly evident.  

\begin{figure}[t!]
	\centering
	\begin{tabular}[b]{@{}cc@{}}
		{\bf (a) Static TacWhisker} & \\
		\hspace{2em}{\bf dabbing motion} & \\
		\includegraphics[width=0.47\columnwidth,trim={0 0 0 0},clip]{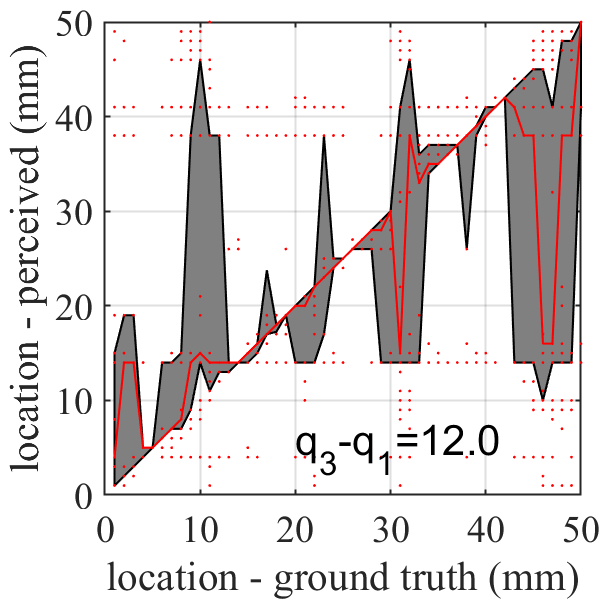} & \\
		{\bf (b) Dynamic TacWhisker} &
		{\bf (c) Dynamic TacWhisker} \\
		\hspace{2em}{\bf whisking motion} &
		\hspace{2em}{\bf calibrated self-motion} \\
		\includegraphics[width=0.47\columnwidth,trim={0 0 0 0},clip]{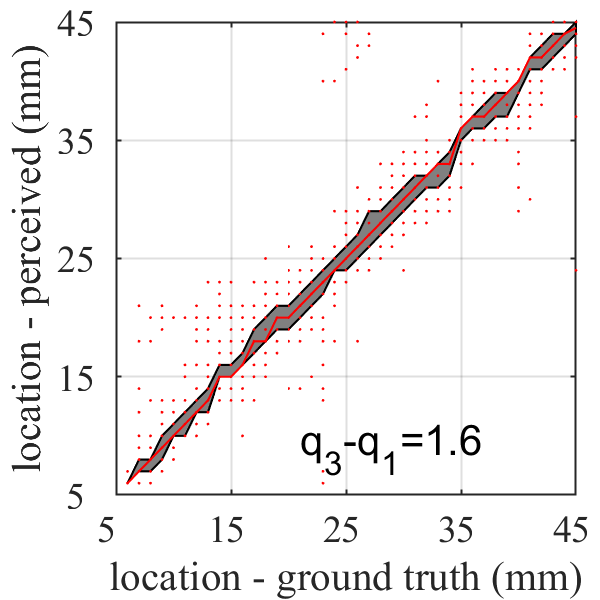} &
		\includegraphics[width=0.47\columnwidth,trim={0 0 0 0},clip]{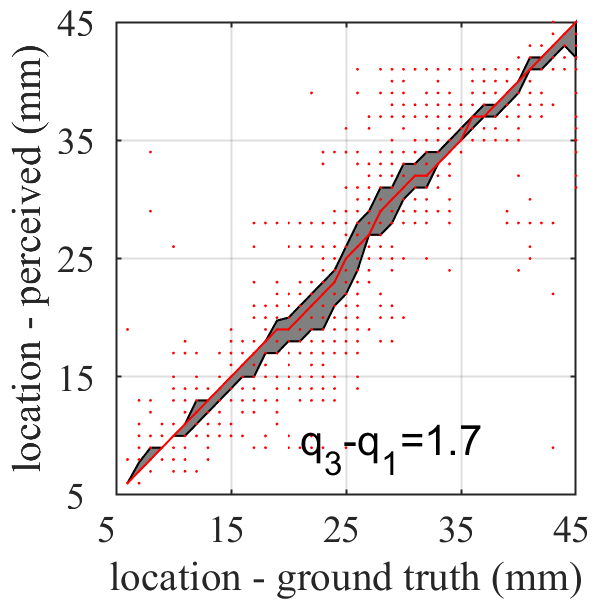} \\
	\end{tabular}
	\caption{Accuracy of location perception for the static (a) and dynamic (c,d) TacWhisker arrays. Monte Carlo 10-fold cross validation (10000 samples) is shown by plotting the the perceived against ground truth locations (red markers). The variability of the location perception is shown between 25th and 75th percentiles (gray region).}
	\label{fig:9}
	\vspace{0em}
\end{figure}

\subsection{Tactile perception depends on TacWhisker motion}
\label{sec:4c}

The accuracy of location perception in the three experimental conditions was then assessed with a standard classifier of tactile data based on a histogram likelihood model (details in Sec.~\ref{sec:3c}). Monte Carlo cross validation over 10 repeated runs was used to generate distributions of the perceived class label against the ground truth class label (Fig.~\ref{fig:9}). 

For the static TacWhisker array (Fig.~\ref{fig:9}a), the location perception is variable for the dabbing motion (interquartile range IQR\,$=12$\,mm). This agrees with qualitative inspection of the data, which on has a sparse, unstructured structure as the rod hits or misses vibrissae (Fig.~\ref{fig:8}a). 

For the dynamic TacWhisker array (Figs~\ref{fig:9}b,c), the location perception is accurate with little variability (interquartile range, IQR\,$=1.5$\,mm), and similar results for the uncompensated and self-motion compensated tactile data. It seems a basic compensation of self-motion does not help the perception, although more sophisticated methods ({\em e.g.} adaptive noise cancellation~\cite{Anderson2010}) may be better. The compensation does reveal a significant covariation of the tactile data with location (Fig.~\ref{fig:8}c), consistent with the accurate perception.

\begin{figure}[t!]
	\centering
	\begin{tabular}[b]{@{}cc@{}}
		{\bf (a) Static TacWhisker} & 
		{\bf (b) Dynamic TacWhisker} \\
		\hspace{2em}{\bf dabbing motion} & 
		\hspace{2em}{\bf whisking motion} \\
		\includegraphics[width=0.47\columnwidth,trim={0 0 0 0},clip]{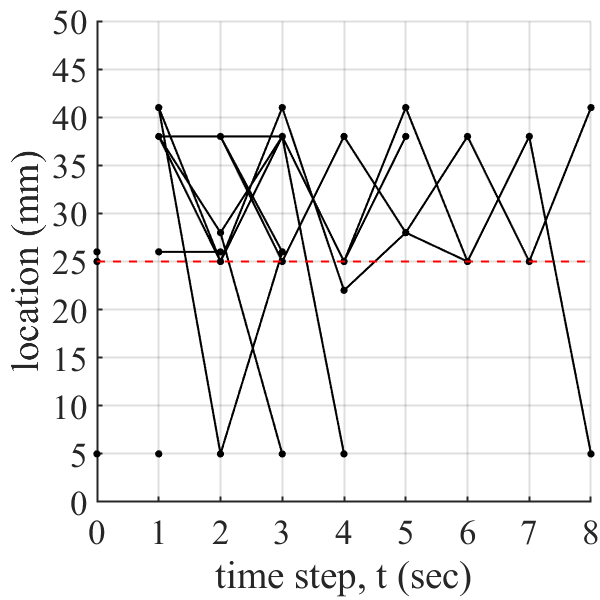} & 
		\includegraphics[width=0.47\columnwidth,trim={0 0 0 0},clip]{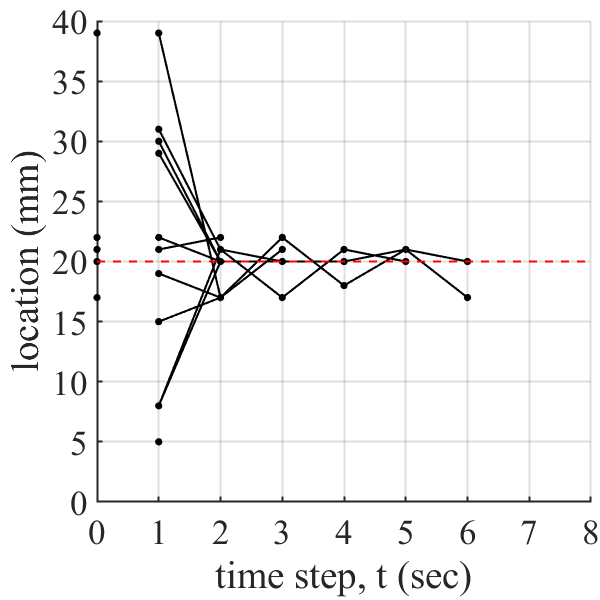} \\
	\end{tabular}
	\caption{Trajectories for actively localizing a stimulus with TacWhisker sensor for the static (a) and dynamic (b) arrays. Trajectories begin from random locations and aim towards the central location (dashed red line). In both cases, a posterior threshold $\theta=0.5$ was used to make a decision.}
	\label{fig:10}
	\vspace{0.5em}
	\centering
	\begin{tabular}[b]{c}
		{\bf Dynamic TacWhisker}\\
		{\bf Location error (active vs passive)}\\
		\includegraphics[width=0.6\columnwidth,trim={0 0 0 0},clip]{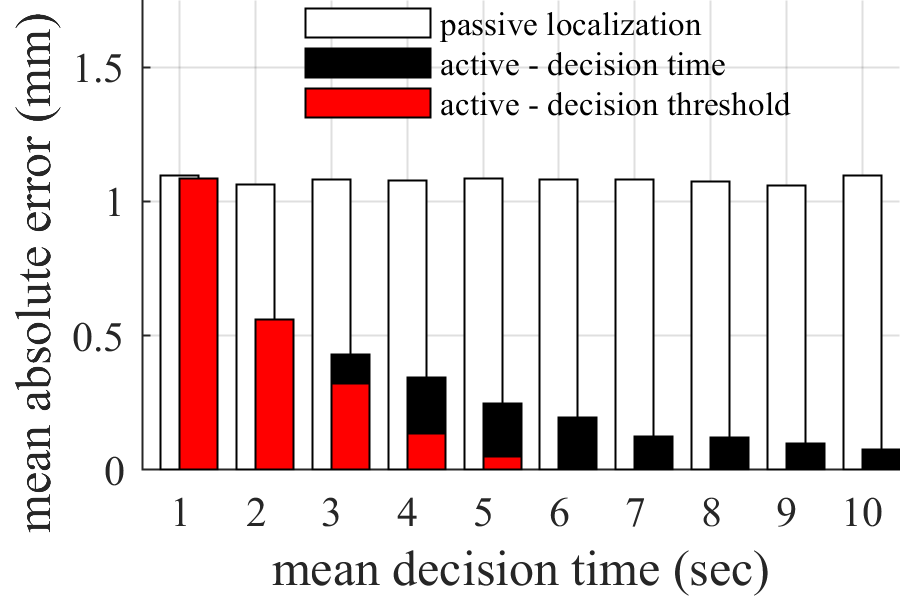}
	\end{tabular}
	\caption{Mean location errors for active and passive touch. Active localization is over multiple contacts while centring on the object, with either a posterior threshold (range 0-0.95; red histogram) or set decision time (1-10 contacts; black histogram). Passive localization does not move the sensor. Averages are over 1000 runs with random starting locations.}
	\label{fig:11}
	\vspace{-1em}
\end{figure}

\subsection{The dynamic TacWhisker aids active touch}
\label{sec:4d}

The location perception was then applied to a simple task in which the TacWhisker actively localizes a stimulus while using intermediate estimates of the object location to centre it within the whisker array (Sec~\ref{sec:3c}). This task is a simple example of active touch~\cite{Lepora2016} and bears analogy with rodent behaviour when exploring stimuli.

For the static TacWhisker array (Fig.~\ref{fig:10}a), the trajectories do not converge on the central location. It appears the quality of the perception from a dabbing motion is not sufficient to actively localize the stimulus within the whisker field.

The dynamic TacWhisker array (Fig.~\ref{fig:10}b) achieves successful active localization across the range of starting locations. All trajectories converge on the central location. 

Repeating over many trials, the mean active localization errors improve with mean decision time (Fig.~\ref{fig:11}, red histogram), reaching near perfect accuracy after $\sim$5 contacts with the threshold-crossing decision rule. A fixed-time rule also improves with decision time but is not as accurate for longer decision times (black histogram). Conversely, mean errors for passive perception do not improve because the robot cannot move to gather new data (white histogram).

Overall, the dynamic TacWhisker array with active localization using a posterior threshold-crossing decision rule gives the best localization performance.

\section{Discussion}

In this paper, we introduced two novel whiskered robots: the static and dynamic TacWhisker arrays (Figs~\ref{fig:4},\ref{fig:5}). These robots are biomimetic in being based on the sensory transduction principles of biological vibrissae, bearing analogy with mechanoreception by Merkel cells in the whisker follicle. We based the designs on a 3D-printed cutaneous (fingertip) optical tactile sensor called the TacTip~\cite{Chorley2009,Ward-Cherrier2018}, which is based on sensory transduction in skin via Merkel cell mechanoreceptors.  The static TacWhisker array modifies the TacTip skin to house 21 whiskers arranged around its tip. The dynamic TacWhisker array is actuated to move its whiskers back and forth in a whisking motion, with the whiskers arranged bilaterally in 2 rows of 5 whiskers.

The performance of the TacWhisker sensors was examined by perceiving the location of a rod, motivated by similar experiment quantifying rodent perception~\cite{Diamond2008}. The quality of the perception depended strongly on the whisker motion. For the static TacWhisker array, a forward dabbing motion was inaccurate and variable (IQR$\sim$10\,mm), consistent with sparse, unstructured contacts.  For the dynamic TacWhisker array, the whisking motion resulted in accurate and reliable perception (IQR$\sim$1.5\,mm).

In consequence, only the dynamic TacWhisker array could perform an active localization task in which stimulus location is estimated while centring the object in the whisker array. Performance improved from a mean error of $\sim$1\,mm after 1 contact to perfect performance after 5 contacts, in accordance with previous studies of active touch~\cite{Lepora2016}. Other active localization strategies could work better for the static TacWhisker (e.g. avoiding locations where the sensing is bad); this would however require a more complex policy for moving the sensor that would need learning. It is also possible that other exploratory motions for the static TacWhisker would improve performance; however, active localization would be far more complex if the contacts were not discrete and independent, as provided by a dabbing or whisking motion.

A potential issue is that the dynamic TacWhisker signal is dominated by the self-generated deflection of the whiskers. This can be partially compensated by subtracting a reference signal (here taken from the central contact); although this did not affect perception, it did improve the visible interpretation of the data. This effect and compensation mechanism are known from animal investigations and have also been considered with the SCRATCHbot whiskered robot~\cite{Anderson2010}. Note that the TacWhisker has greater self-generated signal than both animal and SCRATCHbot whiskers, since the dynamic TacWhisker is directly affected by its actuation (and not just the inertia of its whiskers).

Overall, the TacWhisker arrays give a new class of optical tactile whiskered robots, related to optical tactile sensors that have progressed in recent years~\cite{Ward-Cherrier2018}. They have the benefit of being relatively inexpensive, readily customizable and easy fabrication using multi-material 3D-printing. We expect they can be applied to some of the application domains of tactile whiskers, such as proximity and flow sensing. Furthermore, the biomimetic basis for the TacWhiskers lends them to neurorobotic investigations of rodent tactile vibrissal sensing as an embodied model of animal perception. 

{\em Acknowledgements:} We thank Niels Burnus, Yilin Tao, and members of the Tactile Robotics group: Ben Ward-Cherrier, Nick Pestell, Kirsty Aquilina, Jasper James and John Lloyd. 

%\addtolength{\textheight}{-16cm}   % This command serves to balance the column lengths on the last page of the document manually. It shortens the textheight of the last page by a suitable amount. This command does not take effect until the next page so it should come on the page before the last.

%%%%%%%%%%%%%%%%%%%%%%%%%%%%%%%%%%%%%%%%%%%%%%%%%%%%%%%%%%%%%%%%%%%%%%%%%%%%%%%%

\bibliographystyle{unsrt}
\bibliography{library,Whiskers-robotics}

\end{document}